\documentclass{ws-us}
\usepackage{cite}
\usepackage{url}
\usepackage{makecell}
\usepackage{amsmath}
\usepackage{url}
\usepackage{tabularx}
\usepackage{multicol}
\usepackage[dvipsnames]{xcolor}
\usepackage{soul}

\begin{document}
\setlength{\pdfpagewidth}{8.5in}
\setlength{\pdfpageheight}{11in}
\catchline{0}{0}{2013}{}{}

\markboth{J. Xin et al.}{Real-time Path Planning with SEPSO in Dynamic Scenarios}

\title{Efficient Real-time Path Planning with Self-evolving Particle Swarm Optimization in Dynamic Scenarios}

\author{Jinghao Xin$^{a}$, Zhi Li$^a$, Yang Zhang$^a$, Ning Li$^{a,*}$}

\address{$^a$Department of Automation,
Shanghai Jiao Tong University, Shanghai 200240, P.R. China\\
E-mail: *ning\_li@sjtu.edu.cn}

\maketitle
\begin{abstract}
Particle Swarm Optimization (PSO) has demonstrated efficacy in addressing static path planning problems. Nevertheless, such application on dynamic scenarios has been severely precluded by PSO's low computational efficiency and premature convergence downsides. To address these limitations, we proposed a Tensor Operation Form (TOF) that converts particle-wise manipulations to tensor operations, thereby enhancing computational efficiency. Harnessing the computational advantage of TOF, a variant of PSO, designated as Self-Evolving Particle Swarm Optimization (SEPSO) was developed. The SEPSO is underpinned by a novel Hierarchical Self-Evolving Framework (HSEF) that enables autonomous optimization of its own hyper-parameters to evade premature convergence. Additionally, a Priori Initialization (PI) mechanism and an Auto Truncation (AT) mechanism that substantially elevates the real-time performance of SEPSO on dynamic path planning problems were introduced. Comprehensive experiments on four widely used benchmark optimization functions have been initially conducted to corroborate the validity of SEPSO. Following this, a dynamic simulation environment that encompasses moving start/target points and dynamic/static obstacles was employed to assess the effectiveness of SEPSO on the dynamic path planning problem. Simulation results exhibit that the proposed SEPSO is capable of generating superior paths with considerably better real-time performance (67 path planning computations per second in a regular desktop computer) in contrast to alternative methods. The code and video of this paper can be accessed here\footnote{Code and Video: https://github.com/XinJingHao/Real-time-Path-planning-with-SEPSO}.
\end{abstract}


\begin{multicols}{2}
\section{Introduction}

Path planning, a class of non-deterministic polynomial-time (NP) hard problems, is the computational process of determining a sequence of waypoints for an agent to navigate through an environment from a starting point to a target point while avoiding obstacles and adhering to specific constraints\cite{PPUS}. Path planning serves as a key component of mobile robots that underpins the execution of unmanned missions such as goods distribution, military reconnaissance, and disaster rescue. Dijkstra's algorithm\cite{Djsk} and A*\cite{Astar} are recognized as two canonical and widely utilized path planning algorithms. Nevertheless, despite their extensive adoption, one nonnegligible weakness has severely impaired their real-time performance, restricting them predominantly to the static path planning domain. This weakness arises from the intrinsic sequential computing nature (point-wise sequential manipulation) involved in planning from the start point to the target point, leading to a surge in computational complexity with increasing map complexity (when employing graph-based map representation) or map size (when employing grid-based map representation).

Particle Swarm Optimization\cite{PSO1995}, a population-based metaheuristic optimization technique, has exhibited potentiality as an alternative solution to the aforementioned issue. In PSO-based path planning, the path (a collection of waypoints) is represented by a particle and is evaluated holistically by a predefined fitness function, thereby eschewing the sequential computing of Dijkstra's algorithm and A*. Furthermore, due to its simple concept and easy implementation, PSO has become one of the most popular path planning techniques\cite{ppreview}. However, the standard PSO suffers from inherent flaws, notably premature convergence (due to the loss of population divergence) and low computational efficiency (due to the particle-wise manipulation). Numerous methods have been proposed to bolster the performance of standard PSO during the past decades\cite{DPPSO,IPSO1,IPSO2,IPSO3}. Among these methods, the Diversity-based Parallel Particle Swarm Optimization (DPPSO)\cite{DPPSO} stands out as one of the most recent and promising approaches. DPPSO divides the population into different groups and applies them with diversified hyper-parameters to foster population diversity and mitigate premature convergence. Additionally, the computation of different groups is distributed across individual CPU cores to enable parallel computation and reduce convergence time. Although DPPSO has ameliorated the two weaknesses of standard PSO to some extent, there remain two unresolved issues that demand attention. Firstly, despite DPPSO expediting convergence by resorting to multiprocessing, the low computational efficiency arising from the particle-wise manipulation still exists. Secondly, the diversified hyper-parameters in DPPSO lack a systematic and standardized determination approach. Consequently, we contend further investigation is needed to fill the niche of previous research. 

To this end, we endeavor to tackle the two unresolved issues of DPPSO and develop a more efficient optimization technique to address the real-time path planning problem in dynamic scenarios. The contributions of this paper can be summarized as follows:
\begin{itemize}
\item A TOF for DPPSO is proposed. The TOF converts particle-wise manipulation to tensor operation, which eliminates the coupling between DPPSO's computational complexity and population size. Based on the TOF, we developed a refinement of DPPSO, denoted as Diversity-based Tensor Particle Swarm Optimization (DTPSO). In contrast to the DPPSO, the DTPSO reduces the overall running time on four widely used benchmark functions by 55.3\%(on CPU) and 98.3\%(on GPU) without compromising optimization performance.

\item Leveraging DTPSO's rapidity, we introduce a novel HSEF that autonomously optimizes the diversified hyper-parameters of DTPSO, thereby creating a new variant designated as SEPSO. Experiments on both benchmarks and dynamic path planning problem demonstrate the superiority of SEPSO over DTPSO, DPPSO, and PSO in terms of the final performance, while retaining remarkably better convergence speed.

\item A PI mechanism and an AT mechanism that boost the SEPSO's real-time capabilities on dynamic path planning problem are proposed. Simulation experiments indicate that by incorporating PI and AT, the SEPSO achieves an impressive capability of generating collision-free path in dynamic scenarios at an average frequency of 67 per second. We believe this achievement provides a reassuring solution for addressing real-time path planning problems in dynamic scenarios.

\end{itemize}

The remainder of the paper is structured as follows: An overview of the related works involving PSO, DPPSO, and PSO-based path planning algorithms is introduced in Section 2. Proposed methodologies comprising TOF, HSEF, PI, and AT are elaborated in Section 3. Experimental results and corresponding analysis are presented in Section 4. Conclusions and prospects for future research are delineated in Section 5.

\section{Related Works}

\subsection{Particle Swarm Optimization}
PSO \cite{PSO1995} is a population-based non-gradient stochastic optimization algorithm inspired by the social behavior of birds and fish, which has been wildly leveraged to resolve complex optimization problems, comprising power dispatch\cite{DPPSO}, path planning\cite{PSOpp},  localization\cite{PSOlocation}, etc. 

In PSO, every particle is a representation of a potential solution to the optimization problem and is denoted as $X_{i}=\left(X_{i1}, X_{i2}, \ldots, X_{iD}\right)$, where $ i=1, 2, \ldots, M$ is the particle's index within its population $M$, and $D$ indicates the dimension of the optimization problem. During each iteration, the historical best position of the $i$-th particle, assessed by a predetermined fitness function, is recorded and denoted as $Pbest _{i}=\left(x_{i1}^{\text {Pbest}}, x_{i2}^{\text {Pbest}}, \ldots, x_{\text {iD }}^{\text {Pbest }}\right)$. In addition, the historical best-performing particle among the entire population is identified as the global best and is represented as $Gbest=\left(x_{1}^{\text {Gbest}}, x_{2}^{\text {Gbest}}, \ldots, x_{\text {D}}^{\text {Gbest}}\right)$. The particles adjust their individual velocity $V_i$ and position $X_i$ iteratively based on their own experience and the experience of their neighboring particles: 

\begin{equation}
\begin{aligned}
\label{PSO_V}
V_{i}^{k+1}=\omega^{k} V_{i}^{k}&+C1 \cdot R1^k_{i}  \cdot\left(\mathrm{Pbest}_{i}^{k}-X_{i}^{k}\right) \\
&+C2 \cdot R2^k_{i} \cdot\left(\mathrm{Gbest}^{k}-X_{i}^{k}\right)
\end{aligned}		
\end{equation}

\begin{equation}	
\label{PSO_X}
X_{i}^{k+1}=X_{i}^{k}+V_{i}^{k+1}	
\end{equation}
where $k$ is the iteration counter;
$ C1 $ and $ C2 $  are acceleration constants; 
$ R1^k_{i} $ and $ R2^k_{i} $ are uniform random variables between 0 and 1;
$ \omega^{k} $ is a linear decreasing inertia weight factor: 
\begin{equation}
\label{omega}
\omega^{k}=\omega_{init }-\frac{\omega_{init }-\omega_{end }}{T} \cdot k
\end{equation}
where $T$ is the total number of iterations.
Through this iterative process, the swarm dynamically explores the search space and converges toward the potential optimal solution.

\subsection{Diversity-based Parallel Particle Swarm Optimization}
PSO is notorious for its issues with premature convergence and low computational efficiency. The premature convergence predominantly arises from the loss of population divergence during the search process, while the low computational efficiency can be attributed to the high time complexity resulting from particle-wise manipulation. To mitigate these two issues, \cite{DPPSO} proposed the DPPSO, wherein the population is partitioned into multiple groups, each employing distinct searching parameters. The information among the groups is shared through an asynchronous information sharing mechanism. Such settings facilitate diversified exploration and exploitation capabilities across the groups, thereby alleviating premature convergence. Additionally, the utilization of multiprocessing techniques allows the distribution of different groups' computational tasks to different CPU cores, resulting in a reduction in the overall running time. The updating formula for DPPSO is as follows: 
\begin{equation}
\begin{aligned}
\label{DPPSO_V}
V_{g,n}^{k+1}=\omega^{k}_{g} V_{g,n}^{k}&+C1_{g} \cdot R1^k_{g,n}  \cdot\left(\mathrm{Pbest}_{g,n}^{k}-X_{g,n}^{k}\right) \\
&+C2_{g} \cdot R2^k_{g,n} \cdot\left(\mathrm{Gbest}^{k}_{g}-X_{g,n}^{k}\right) \\
&+C3_{g} \cdot R3^k_{g,n} \cdot\left(\mathrm{Tbest}^{k}-X_{g,n}^{k}\right) 
\end{aligned}		
\end{equation}

\begin{equation}	
\label{DPPSO_X}
X_{g,n}^{k+1}=X_{g,n}^{k}+V_{g,n}^{k+1}	
\end{equation}
where $g=1,2,...,G$ is the index of the group that the particle belongs to; 
$n=1,2,...,N$ is the index of the particle within its group;  
$ C1_{g}, C2_{g}, C3_{g} $  are the acceleration constants; 
$ R1^k_{g,n}, R2^k_{g,n}, R3^k_{g,n} $ are uniform random variables between 0 and 1; 
$\mathrm{Pbest}_{g,n}^{k}$ is the best position found by particle $[g, n]$ within iteration $k$;
$\mathrm{Gbest}^{k}_{g}$ is the best position identified by group $g$ within iteration $k$;
$\mathrm{Tbest}^{k}$ is the best position discovered by the whole population within iteration $k$.

Although DPPSO has mitigated the premature convergence and low computational efficiency issues of PSO, there still remains two notable unresolved concerns. Firstly, while the authors of DPPSO circumvent the prolonged running time by employing multiprocessing, the high time complexity flaw inherent in both PSO and DPPSO persists. In light of this, we propose the TOF, aimed at reducing the time complexity of DPPSO and enhancing execution speed from the foundational level. Secondly, the authors of DPPSO configured its hyper-parameters based on expert knowledge. Nevertheless, it remains uncertain whether these hyper-parameters are suitable for problems in other domains. Consequently, we introduce the HSEF to systematically determine diverse hyper-parameters.

\subsection{Real-time Path Planning with PSO}
Lai et al.\cite{BSCP} have developed a real-time local motion planning algorithm based on PSO. In \cite{BSCP}, the trajectory is encoded by motion primitives and is optimized by PSO in a model predictive control fashion. Zhou et al.\cite{formation} have extended Lai's work\cite{BSCP} to a multi-agent case. Despite their successful application on real-world rotorcraft motion planning, there are two aspects in their algorithms that necessitate refinement: (i) Optimization technique. Both \cite{BSCP} and \cite{formation} employed raw PSO as the optimization technique. However, raw PSO bears premature convergence and low computational efficiency downsides. (ii) Map representation. To realize collision detection, \cite{BSCP} and \cite{formation} have resorted to occupied grid map or Euclidean distance transform map. Nevertheless, the computational complexity of these two maps can escalate expeditiously with the growth of map size or resolution. In light of this, we believe further work is needed to unleash the maximum potential of PSO-based path planning algorithms.

\section{Methodology}

In this section, we commence by introducing the TOF and presenting the DTPSO. Subsequently, leveraging the computational bonus of DTPSO, we proceed to elucidate the HSEF and then propose the SEPSO. Finally, we demonstrate how to use the SEPSO to efficiently address path planning in dynamic scenarios with the assistance of the PI and AT mechanisms.

\subsection{TOF and DTPSO}
The DPPSO relies on the iteration formulated by Eq.~(\ref{omega}), (\ref{DPPSO_V}), (\ref{DPPSO_X}). Although such expression is straightforward, it is computationally inefficient due to its particle-wise manipulation. That is, the velocity, position, and inertia factor are updated individually for each particle, giving rise to the interconnection between the population size and time complexity. To surmount this limitation, we assemble similar components into tensors and propose the TOF for DPPSO, represented by Eq.~(\ref{TOF_omega}), (\ref{TOF_V}), (\ref{TOF_X}).
\begin{equation}
\label{TOF_omega}
\overrightarrow{\omega^k}=[\omega_1^k, \omega_2^k, \cdots,\omega_G^k]=\overrightarrow{\omega^{init}}-\frac{\overrightarrow{\omega^{init}}-\overrightarrow{\omega^{end}}}{T} k
\end{equation}
\begin{equation}
\label{TOF_V}
V^{k+1}=I \times[H \cdot R \cdot(K-L)]
\end{equation}
\begin{equation}	
\label{TOF_X}
X^{k+1}=X^{k}+V^{k+1}	
\end{equation}
where `$\times$' is matrix product; `$\cdot$' is element-wise product;

$H = \left[\begin{array}{cccc}
\omega_1^k & \omega_2^k & \cdots & \omega_G^k \\
C 1_1 & C 1_2 & \cdots & C 1_G \\
C 2_1 & C 2_2 & \cdots & C 2_G \\
C 3_1 & C 3_2 & \cdots & C 3_G
\end{array}\right]$; 
$R=\left[\begin{array}{cccc}
\overrightarrow{1} & \overrightarrow{1} & \cdots & \overrightarrow{1} \\
R 1_1^k & R 1_2^k & \cdots & R 1_G^k \\
R 2_1^k & R 2_2^k & \cdots & R 2_G^k \\
R 3_1^k & R 3_2^k & \cdots & R 3_G^k
\end{array}\right] $;\\

$K=\left[\begin{array}{cccc}
V_1^k & V_2^k & \cdots & V_G^k \\
P_1^k & P_2^k & \cdots & P_G^k \\
G_1^k & G_2^k & \cdots & G_G^k \\
T^{\mathrm{k}} & T^{\mathrm{k}} & \cdots & T^{\mathrm{k}}
\end{array}\right]$;
$L=\left[\begin{array}{cccc}
\overrightarrow{0} & \overrightarrow{0} & \cdots & \overrightarrow{0} \\
X_1^k & X_2^k & \cdots & X_G^k \\
X_1^k & X_2^k & \cdots & X_G^k \\
X_1^k & X_2^k & \cdots & X_G^k
\end{array}\right]$

Please refer to Table \ref{EXP_TOF} for the definition and explanation of the symbols in the TOF. The TOF accomplishes the calculation of Eq.~(\ref{omega}), (\ref{DPPSO_V}), (\ref{DPPSO_X}) in a fashion of tensor operation, disentangling the interconnection between the population size and time complexity. This approach bolsters computational efficiency and leads to significant reductions in the overall running time. Furthermore, this advantage could be further exaggerated when deployed on hardware platforms that are optimized for tensor operations, such as Graphics Processing Unit (GPU) and Tensor Processing Unit (TPU). Based on the TOF, we put forward a refinement of the DPPSO, denoted as DTPSO, as given in Algorithm 1. Note that the hyper-parameters of DTPSO that controls the searching behavior (balance between exploration and exploitation) are inherited from DPPSO and are a matrix of shape $(G,6)$ as listed in Table \ref{hyper8}.

\begin{tablehere}
\tbl{Explanation of the Symbol in TOF. \label{EXP_TOF}}
{\begin{tabular}{lll}
\toprule
Symbol   & Dimension   & Explanation\\
\colrule
$I$ & (1,4) & [1, 1, 1, 1]\\
$H$ & (4,G) & Hyper tensor\\	
$R$ & (4,G,N) & Random tensor\\	
$K$ & (4,G,N,D)& Kinematics tensor\\	
$L$ & (4,G,N,D)& Location tensor\\	
$V^k$& (G,N,D) & Velocity tensor at iteration k\\
$X^k$& (G,N,D) & Position tensor at iteration k\\
$\overrightarrow{\omega^{init}}$& (1,G) & $[\omega_1^{init}, \omega_2^{init}, \cdots ,\omega_G^{init}]$\\	
$\overrightarrow{\omega^{end}}$& (1,G) & $[\omega_1^{end}, \omega_2^{end}, \cdots ,\omega_G^{end}]$\\	
$\omega^k_{g}$& (1,)& \makecell[l]{Inertia factor of group $g$\\at iteration $k$}\\	
$C1_g,C2_g,C3_g$ & (1,) & \makecell[l]{Acceleration constants of group $g$}\\
$\overrightarrow{1}$& (N,)& Constant tensor 1\\ 
$R1_g^k,R2_g^k,R3_g^k$ & (N,) & \makecell[l]{Random variables of group $g$ at\\ iteration $k$; $N$ distinct scalars}\\
$V_g^k$ & (N,D) & \makecell[l]{Velocity tensor of group $g$\\at iteration $k$} \\	
$P_g^k$ & (N,D) & \makecell[l]{$Pbest$ tensor of group $g$\\at iteration $k$} \\	
$G_g^k$ & (N,D) & \makecell[l]{$Gbest$ tensor of group $g$\\at iteration $k$} \\
$T^k$ & (N,D) & \makecell[l]{$Tbest$ tensor at iteration $k$} \\
$\overrightarrow{0}$& (N,D)& Constant tensor 0\\ 
$X_g^k$ & (N,D) & \makecell[l]{Position tensor of group $g$\\at iteration $k$} \\
		
\botrule
\multicolumn{3}{l}{
\makecell[l]{Note that every row of $G_g^k$ is identical, and every row of $T^k$ is \\
 also identical.}}
\end{tabular}}
\end{tablehere}

\begin{tablehere}
\label{algo:1}
{
\resizebox{0.49\textwidth}{!}{
\begin{tabular}{l}
\toprule
\textbf{Algorithm 1}: Diversity-based Tensor Particle Swarm Optimization\\
\colrule
Initialize the particles and hyper-parameters, and let $k=0$.\\
\textbf{while} k $<$ T:\\
\quad k$\leftarrow$k+1 \\
\quad Calculate fitness value for all particles: $Fitness$ = F($X^k$) \\
\quad Update $P^k_g$, $G^k_g$, $T^k$ when better $Fitness$ is identified\\
\quad Update $\overrightarrow{\omega^k}$ according to Eq.~(\ref{TOF_omega}) \\
\quad Update $V^k$ according to Eq.~(\ref{TOF_V}), and clip $V^k$ to legal interval $V_{range}$ \\
\quad Update $X^k$ according to Eq.~(\ref{TOF_X}). and clip $X^k$ to legal interval $X_{range}$ \\

\textbf{return} F($T^k$) as optimum\\

\botrule
\end{tabular}}}
\end{tablehere}

\begin{figurehere}
\begin{center}
\centerline{\includegraphics[width=0.5\textwidth]{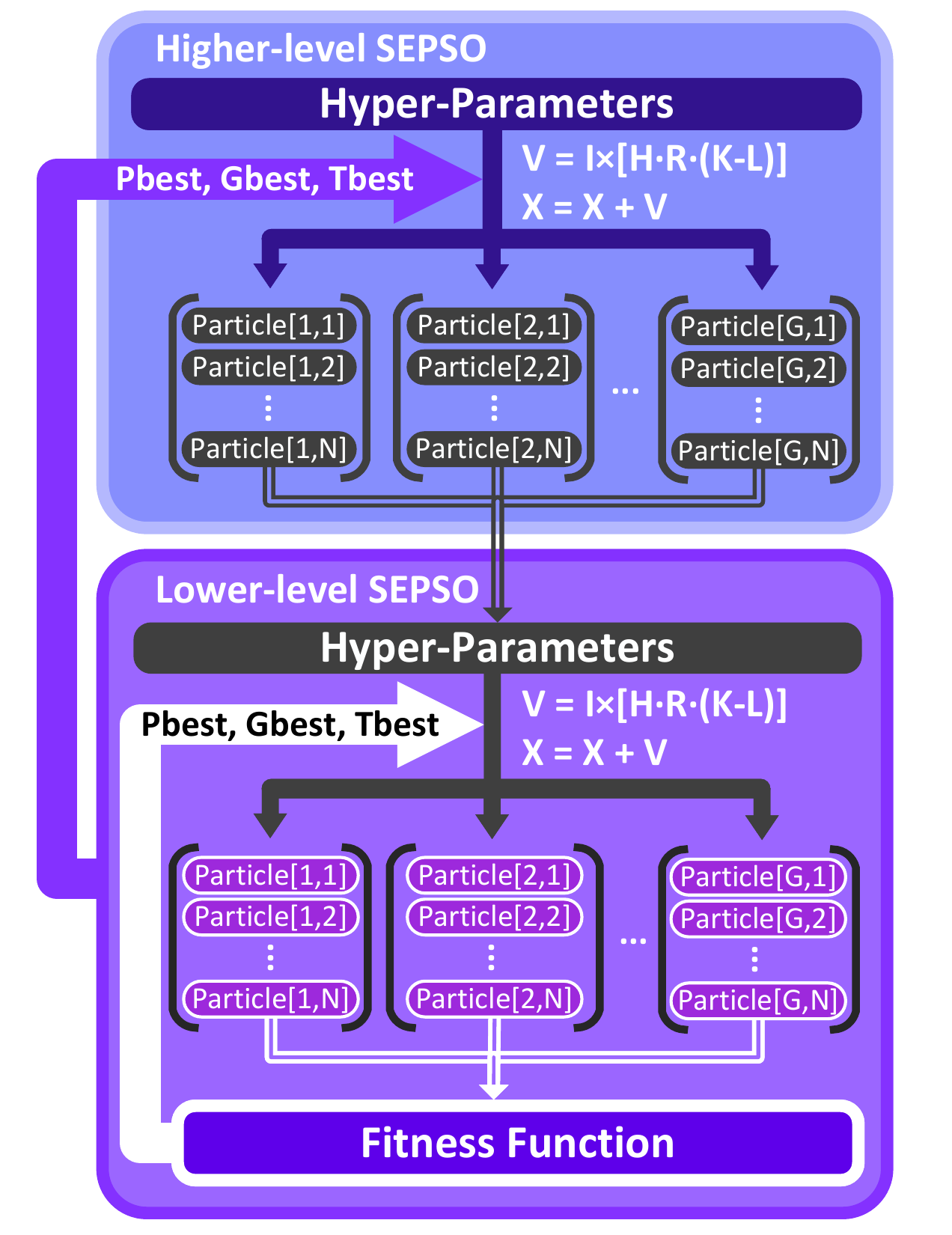}}
\caption{\textcolor{black}{Hierarchical Self-evolving Framework.}}
\label{HSEF}
\end{center}
\end{figurehere}

\subsection{HSEF and SEPSO}
\subsubsection{Hierarchical Self-evolving Framework}\label{3.2}

An unresolved matter in both DPPSO and the proposed DTPSO pertains to the determination of hyper-parameters. To this end, we propose the SEPSO, which is underpropped by a HSEF that enables autonomous tunning of its own hyper-parameters, as illustrated by Fig. \ref{HSEF}. The HSEF comprises two distinct components: the Higher-level Self-Evolving Particle Optimization (H-SEPSO) and the Lower-level Self-Evolving Particle Optimization (L-SEPSO). Both H-SEPSO and L-SEPSO are underpinned by the DTPSO introduced in the preceding section, with minor variations in the definition of fitness functions and particles. Specifically, the L-SEPSO corresponds directly to the DTPSO, where the fitness function aligns with the specific optimization problem under consideration, and each particle represents a potential solution to this problem. The L-SEPSO accepts one set of hyper-parameters as its input and maps it to the Lowest Fitness Value (LFV) attainable after a fixed number of iterations $T$:
\begin{equation}
\label{fit_h}
LFV = \text{L-SEPSO}(\text{hyper-parameters})
\end{equation}

Accordingly, the particles employed in the H-SEPSO represent diverse sets of prospective hyper-parameters for the L-SEPSO and are evaluated by the fitness function formulated by Eq.~(\ref{fit_h}). Through the iteration of the H-SEPSO, better hyper-parameters for the L-SEPSO could be found and a stronger optimization performance for L-SEPSO could be promised. We define the entire process as the evolution phase of the SEPSO. Conventionally, the evolution phase is carried out offline. Following the evolution phase, the best-discovered hyper-parameters are extracted and applied in conjunction with the DTPSO during the online optimization process, referred to as the application phase of the SEPSO.

\subsubsection{Bootstrap}
One might wonder whether it is possible to evolve by bootstrap. That is, reload the hyper-parameters of the H-SEPSO with the best hyper-parameters found so far at the onset of every evolution. Although it is appealing to do so, it does not yield favorable experimental outcomes. A reasonable exposition might be the discrepancy between the fitness function of the H-SEPSO and L-SEPSO, such that good hyper-parameters for the L-SEPSO could not generalize to and suit the H-SEPSO. In this context, the H-SEPSO uses the default hyper-parameters recommended by DPPSO\cite{DPPSO} as listed in Table \ref{hyper8}.

\subsubsection{Necessity of DTPSO}
Despite the HSEF is generally compatible with other optimization techniques from the PSO family, its optimal performance is achieved when paired with the proposed DTPSO. The primary reason is that the HSEF can be computation-hungry, as each evaluation of the particles in H-SEPSO corresponds to an entire optimization process of the L-SEPSO, necessitating the computational efficiency design of the L-SEPSO, which is exactly the aim of DTPSO.

\subsection{Path Planning with SEPSO}
\subsubsection{Map representation}
The polygonal map\cite{pmap} is employed to represent the path and obstacles, as depicted in Fig. \ref{path}. Specifically, the path is depicted as a set of waypoints from the starting point to the target point. Meanwhile, the obstacles are represented by the vertexes of their respective bounding boxes. This polygonal representation offers three notable advantages. Firstly, the environmental information can be efficiently stored by recording the coordinates of the waypoints, start and target points, as well as the vertexes of the bounding box, making this representation memory-friendly. Additionally, in contrast to commonly employed occupied grid map\cite{color}, the computational complexity of the polygonal map does not escalate with the expansion of the map size, rendering it feasible to perform real-time path planning on large-scale dynamic maps. Secondly, the problem of collision detection between the path and obstacles can be transformed into a problem of segments intersection involving the path segments and the bounding box segments, as delineated in Fig. \ref{obstype}. This method facilitates the handling of obstacles in any shape, be it convex or non-convex. Thirdly, the transformation enables expeditious and efficient calculation through matrix operations, especially when deployed on GPU or TPU. However, due to the extensive nature of elaborating on the third advantage, which goes beyond the core contribution of this paper, we kindly refer readers to consult our provided code\footnotemark[1] for more comprehensive insights.

\begin{figurehere}
\begin{center}
\centerline{\includegraphics[width=0.37\textwidth]{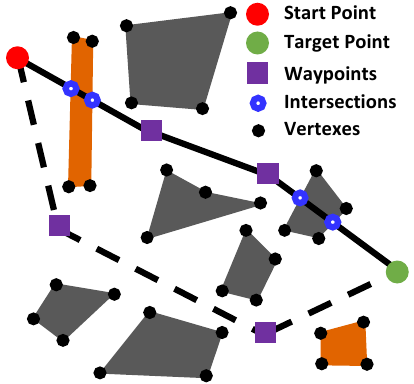}}
\caption{Polygonal representation of path and obstacles. Note that the obstacles in orange and black are static obstacles and dynamic obstacles, respectively.}
\label{path}
\end{center}
\end{figurehere}

\begin{figurehere}
\begin{center}
\centerline{\includegraphics[width=0.48\textwidth]{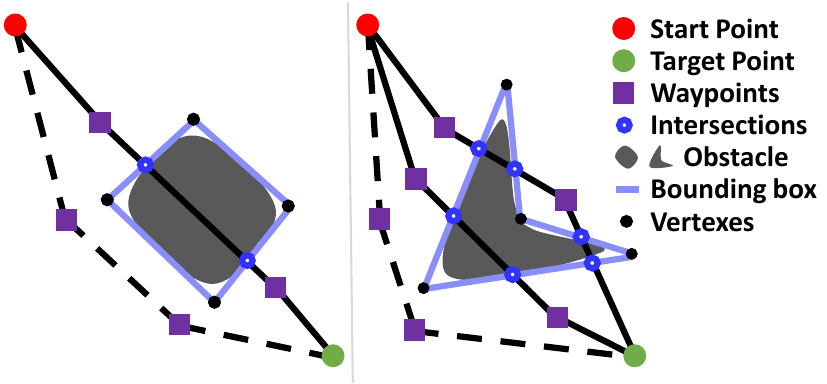}}
\caption{\textcolor{black}{An illustration of collision detection involving convex (left) and non-convex (right) obstacles. Here, the solid paths would be penalized due to their intersection with the bounding box, whereas the dashed paths would not.}}
\label{obstype}
\end{center}
\end{figurehere}

\subsubsection{Definition of Particles and Fitness Function}
Prior to implementing the SEPSO to tackle the path planning problem, it is imperative to establish clear definitions for both ``particle" and ``fitness function". As expounded in Section \ref{3.2}, each particle of the H-SEPSO corresponds to a potential combination of the hyper-parameters for the L-SESPO, and the fitness values of these particles are the LFV mapped by the L-SEPSO. Correspondingly, each particle of the L-SEPSO represents a potential path, and are denoted by
\begin{equation}
\label{coordinate}
X_{g,n}=[x_1,x_2,...,x_{D/2},y_1,y_2,...,y_{D/2}]
\end{equation}
where $x_i$ and $y_i$ are the coordinates of the $i$-th waypoint.

The fitness function of the L-SEPSO for path planning is given by
\begin{equation}
\label{path_fit}
F(X_{g,n})=Length(X_{g,n})+Penalty(X_{g,n})
\end{equation}
where the $Length(X_{g,n})$ designates the length of the path, and the $Penalty(X_{g,n})$ corresponds to the penalty resulting from the collision between the path and the bounding boxes of the obstacles:

\begin{equation}
\begin{aligned}
\label{lenth}
Length(X_{g,n})&=\sum^{D/2}_{d=2}\sqrt{(x_d-x_{d-1})^2+(y_d-y_{d-1})^2}\\
&+\sqrt{(x_s-x_{1})^2+(y_s-y_{1})^2}\\
&+\sqrt{(x_t-x_{D/2})^2+(y_t-y_{D/2})^2}
\end{aligned}
\end{equation}

\begin{equation}
\label{penalty}
Penalty(X_{g,n})=\alpha \cdot Q(X_{g,n})^\beta
\end{equation}
where $(x_s,y_s)$ and $(x_t,y_t)$ are the coordinates of the start and target point, $\alpha$ and $\beta$ are two parameters that control the strength of the penalty, and $Q(X_{g,n})$ is the number of intersections between the path and the bounding boxes of the obstacles. As depicted in Fig. \ref{path},  $Q(\text{Dashed Path})=0$, while $Q(\text{Solid Path})=4$.

\subsubsection{Priori Initialization}
The convergence speed of path planning is of paramount importance in the context of real-time path planning. To expedite the convergence speed of SEPSO at its application phase, we have devised a PI mechanism, as illustrated in Fig. \ref{PI}, to initialize particle positions at the onset of each planning process. Detailly, the PI leverages the planning results from the preceding time step as prior knowledge and initializes the position of the particles within its PI interval. The rationale behind the PI lies in the continuous movement of obstacles and the start/target points, where the optimal paths of two consecutive time steps are expected to exhibit gradual changes rather than abrupt transitions in most case. However, it is essential to acknowledge that PI may lead to suboptimal solutions, as it accelerates the convergence of the swarm at the expense of limiting its exploration capabilities. To maintain a balance between convergence speed and exploration, the PI is merely applied to a fraction, denoted as $\gamma$, of the particles within each group of the SEPSO. Note that the PI is omitted for the initial time step due to the absence of a preceding path.

\begin{figurehere}
\begin{center}
\centerline{\includegraphics[width=0.46\textwidth]{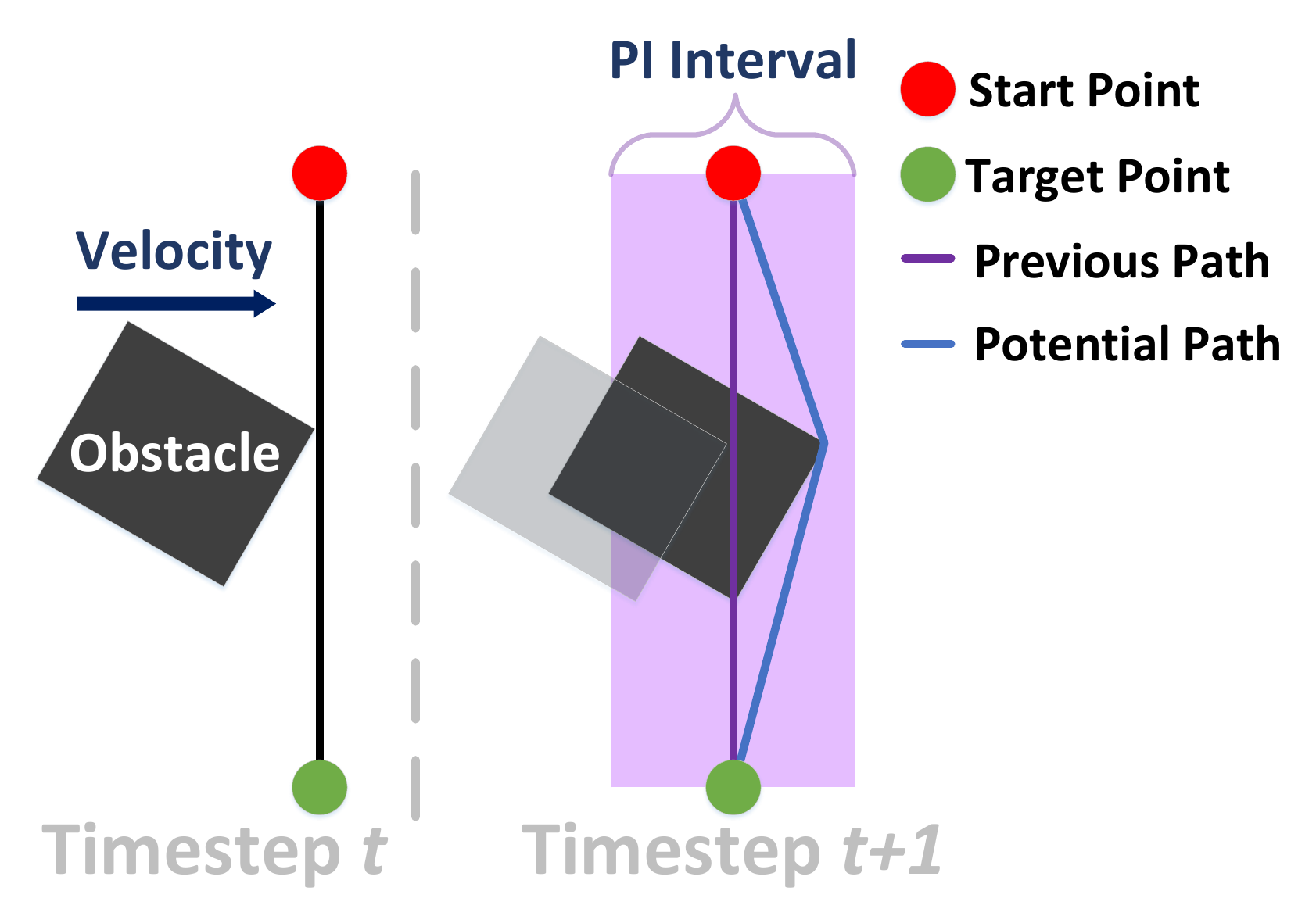}}
\caption{\textcolor{black}{Priori Initialization Mechanism.}}
\label{PI}
\end{center}
\end{figurehere}

\begin{table*}
\tbl{Details of 4 benchmark functions.\label{BFs}}
{\begin{tabular}{@{}lllllll@{}}
		\toprule
		No.&Function Name&Dimension($D$)&Range of search($X_{range}$)&Function Type&Problem Type&Minimum\\
		\colrule
		$BF_1$&Sphere function&30&${[-600,600]}^D$&Unimodal&Minimization&0\\
		$BF_2$&Rosenbrock function&30&${[-600,600]}^D$&Unimodal&Minimization&0\\
		$BF_3$&Rastrigin function&30&${[-600,600]}^D$&Multimodal&Minimization&0\\
		$BF_4$&Griewank function&30&${[-600,600]}^D$&Multimodal&Minimization&0\\
		\botrule
\end{tabular}}
\end{table*}

\begin{figure*}
\begin{center}
\centerline{\includegraphics[width=0.95\textwidth]{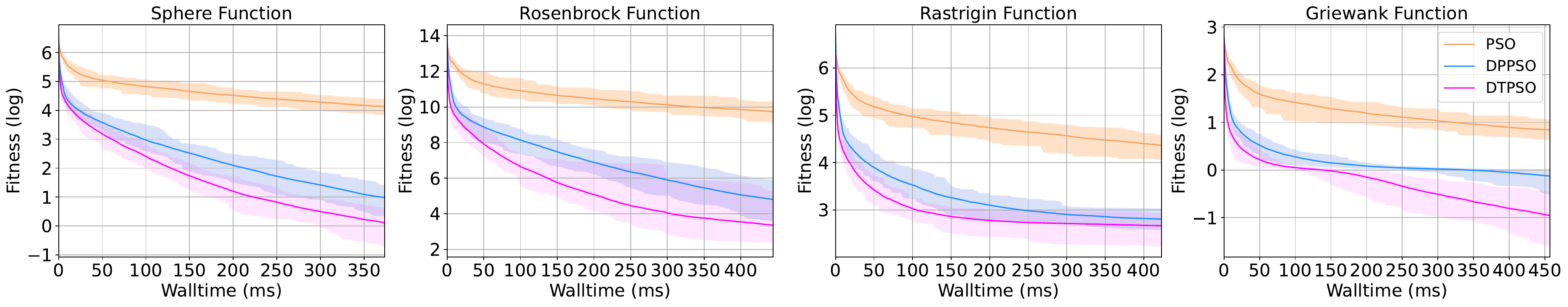}}
\caption{Computational efficiency comparison between DTPSO, DPPSO, and PSO. The solid curves represent the mean value over the 50 trials and the translucent region corresponds to the maximal and minimal range.}
\label{bf_time}
\end{center}
\end{figure*}
\subsubsection{Auto Truncation}
In the context of dynamic path planning, the computational complexity fluctuates in accordance with the movement of obstacles. Employing a fixed iteration number for each planning process is thus inadequate. A small iteration number may lead to unfavorable solution quality, whereas an excessively large iteration number results in increased planning latency. Consequently, we developed the AT mechanism that truncates the iteration of SEPSO at its application phase automatically. The AT calculates the standard deviation of the L-SEPSO's LFV over the last Truncation Window ($TW$) iterations and truncates the iteration when the following two criteria are satisfied:

\begin{itemize}
\item The standard deviation of the L-SEPSO's LFV over the last $TW$ iterations is smaller than a small constant $\delta$.
\item The current best path is collision-free.
\end{itemize}

\section{Experiments}
In this section, we first evaluate the proposed DTPSO and SEPSO on four widely used nonlinear benchmark functions \cite{DPPSO,Izci2020,Rahman2019} to ascertain the effectiveness of the proposed methods. The details of these benchmark functions are given in Table \ref{BFs}. Subsequently, we utilize SEPSO to tackle the path planning problem in dynamic scenarios, and a comparative analysis of the results is conducted with other existing methods. The hardware platform underprops these experiments is reported in Table \ref{EP}.

\subsection{Benchmark Evaluation of DTPSO}
\subsubsection{Standard benchmark functions}\label{4.1.1}
For a fair comparison, we follow the experimental setups from DPPSO\cite{DPPSO}. Specifically, the total number of particles of DTPSO, DPPSO, and PSO is all 80. Note that DTPSO and DPPSO are divided into 8 groups, and each group is configured with diversified hyper-parameters, as given in Table \ref{hyper8}. The total iteration for three algorithms is 1400. To overcome the randomness, the experiments are conducted for 50 independent trials. Afterwards, the computational efficiency of the three algorithms is compared by a measurement of wall-clock time, as illustrated in Fig. \ref{bf_time}. The result reveals that the DTPSO outstrips both DPPSO and PSO within the same timeframe on all benchmark functions, corroborating the superiority in computational efficiency of the TOF.  In addition, for the completeness of the experimental results, we also present the fitness curves by a measurement of iterations and the overall running time of the corresponding algorithms, as respectively presented in Fig. \ref{bf_iteration} and Table \ref{standard_bf}.

\subsubsection{Large-scale benchmark functions}
To further demonstrate the efficiency of TOF on large-scale optimization problems when combined with GPU, we extend the dimension of the 4 benchmark functions and compare the time consumed for 1400 iterations. Detailedly, the number of particles ($N$) is set as 16$K$ and the dimension of the solution ($D$) is set as 1$K$. In this context, both the position and velocity of the particle swarm correspond to 16$M$ scalars. The experiment is repeated for 50 independent trials, and the averaged results are compared in Table \ref{large_bf}. Note that the DTPSO(GPU) is implemented with PyTorch\footnote{https://pytorch.org/}.

\begin{tablehere}
\tbl{Running Time (s) Comparison on Large-scale Benchmark Functions. \ \ \ \ \ \ \ \ \ \ \ \ \ \ \ \ \ \ \ \ \ \ \ \ \ \ \ \label{large_bf}}
{\begin{tabular}{lllll}
    \toprule
    No.   & PSO   & DPPSO & DTPSO(CPU) & DTPSO(GPU) \\
	\colrule
    BF1   & 10487.9  & 1583.2  & 659.8  & 23.8  \\
    BF2   & 10910.9  & 1690.7  & 848.3  & 30.3  \\
    BF3   & 11912.3  & 1802.7  & 774.5  & 29.6  \\
    BF4   & 12025.1  & 1779.7  & 782.0  & 30.2  \\
	\botrule
\end{tabular}}
\end{tablehere}

The findings suggest that in the context of large-scale optimization problems, the DTPSO (CPU) results in a reduction of approximately 55.3\% and 93.2\% in the overall running time compared to DPPSO and PSO, respectively. Furthermore, when deployed with GPU, the DTPSO (GPU) yields a notable reduction of roughly 98.3\% and 99.8\% in the running time when contrasted with DPPSO and PSO, respectively. The obtained results strikingly mirror the significant computational efficiency advantages stemming from the TOF while also revealing the prospective applicability of our approach in handling large-scale optimization problems, such as neural network optimization and hyper-parameter tuning of complex systems.

\subsection{Benchmark Evaluation of SEPSO}
\subsubsection{Evolution Phase on Benchmark Functions}
To validate the effectiveness of the HSEF, we first perform the experiment concerning the evolution phase of SEPSO on 4 benchmark functions. Specifically, the L-SEPSO follows the experimental setups from Section \ref{4.1.1}. The only difference is that the hyper-parameters of the L-SEPSO are inherited from the particles of H-SEPSO at every evolution. Regarding the H-SEPSO, the hyper-parameters remain stationary and are listed in Table \ref{hyper8}. We evolve the SEPSO for 500 evolutions and plot the fitness curves of the H-SEPSO in Fig. \ref{SE_curves}.

Recalling Eq.~(\ref{fit_h}), since the fitness value of the H-SEPSO denotes the LFV obtained by the L-SEPSO, the descending curves in Fig. \ref{SE_curves} illustrate that, as the self-evolution progresses, the H-SEPSO systematically identifies a better combination of hyper-parameters for the L-SEPSO, resulting in the enhanced optimization capabilities of L-SEPSO. 

\begin{figurehere}
\begin{center}
\centerline{\includegraphics[width=0.4\textwidth]{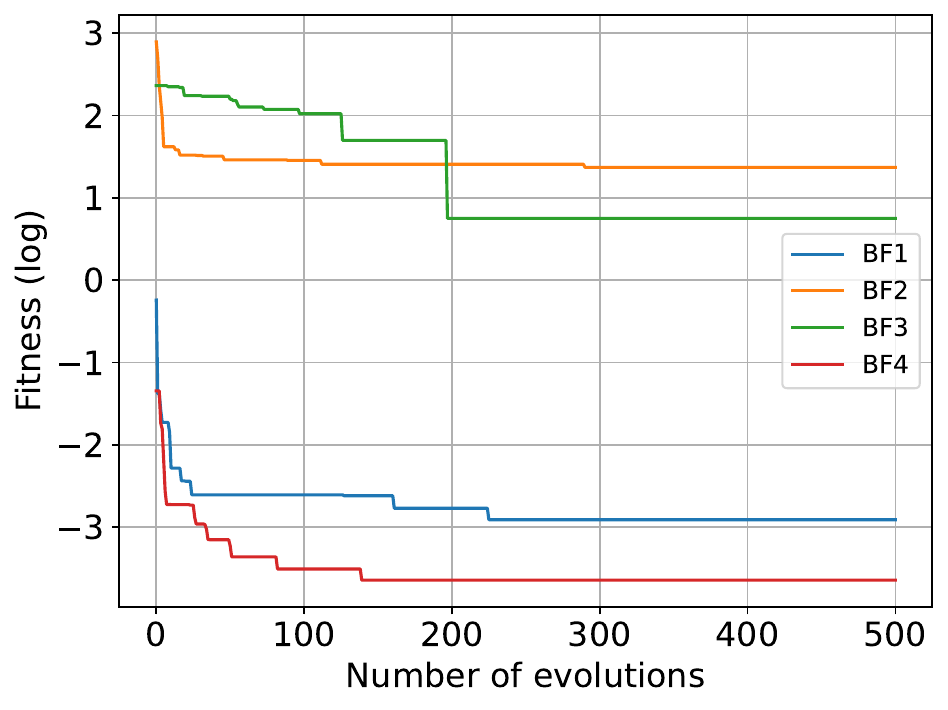}}
\caption{Fitness curves of the Higher-level SEPSO on 4 benchmark functions.}
\label{SE_curves}
\end{center}
\end{figurehere}

\begin{figure*}
\begin{center}
\centerline{\includegraphics[width=0.98\textwidth]{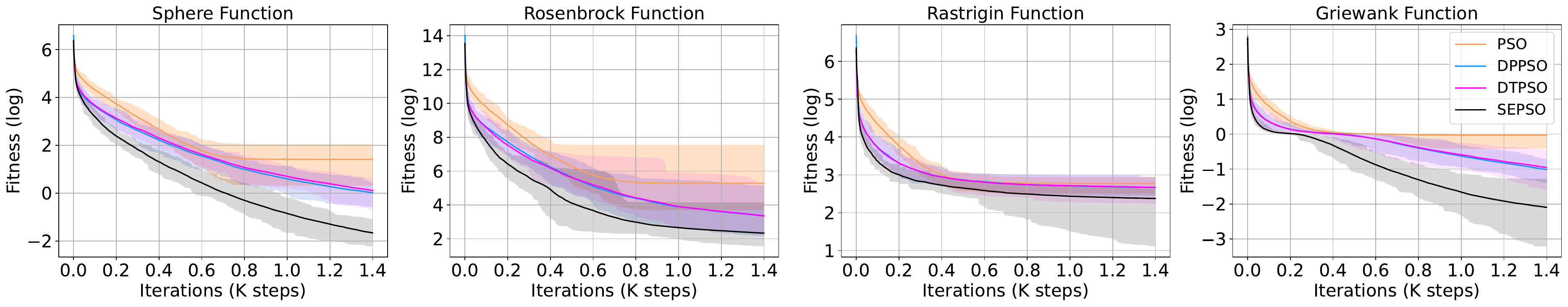}}
\caption{Fitness curves of SEPSO, DTPSO, DPPSO, and PSO on benchmark functions. The solid curves represent the mean value over the 50 trials and the translucent region corresponds to the maximal and minimal range.}
\label{bf_iteration}
\end{center}
\end{figure*}

\begin{figure*}
\begin{center}
\centerline{\includegraphics[width=0.98\textwidth]{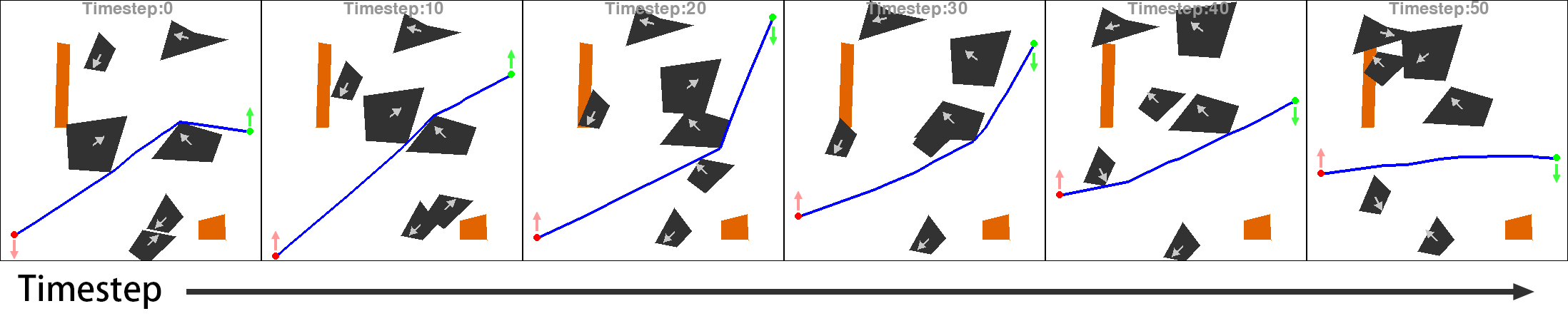}}
\caption{Snapshots of the simulation results at Timestep 0, 10, 20, 30, 40, and 50. The black and orange blocks are the dynamic and static obstacles, respectively. To prevent the dynamic obstacles from exceeding the boundaries of the map, their velocities are reversed when they reach the map's borders. The start point and target point are depicted in red and green, with a respective velocity of $3\ cm/s$ and $8\ cm/s$ moving along the vertical axis. The velocities of the dynamic objects in each timestep are marked with their respective arrows. Note that the video of the simulation results is available on our website\protect\footnotemark[1].}
\label{PP}
\end{center}
\end{figure*}

\subsubsection{Application Phase on Benchmark Functions}
After the evolution phase, we extract the best hyper-parameters for each of the 4 benchmark functions and evaluate them with the DTPSO, following the experimental configurations outlined in Section \ref{4.1.1}. The resulting approach is referred to as the application phase of SEPSO. The fitness curves are presented in Fig. \ref{bf_iteration} and the running time comparison is listed in Table \ref{standard_bf}. The obtained results demonstrate that the HSEF considerably elevates the optimization capabilities of the DTPSO, reaching substantially better solutions within an identical number of iterations. Meanwhile, as observed in the DTPSO, the TOF remarkably enhances the computational efficiency of the SEPSO as well, resulting in respective reductions of the running time by approximately 52.9\% and 92.6\% when compared to DPPSO and PSO among standard benchmark functions.

\begin{tablehere}
\tbl{Running Time (s) Comparison on Standard Benchmark Functions. \ \ \ \ \ \ \ \ \ \ \ \  \label{standard_bf}}
{\resizebox{0.4\textwidth}{!}{
\begin{tabular}{lllll}
    \toprule
    No.   & PSO   & DPPSO & DTPSO & SEPSO\\
	\colrule
    BF1   & 3.98  & 0.65  & 0.37  & 0.37  \\
    BF2   & 5.99  & 0.92  & 0.44  & 0.44  \\
    BF3   & 6.20  & 0.96  & 0.42  & 0.42  \\
    BF4   & 6.91  & 1.08  & 0.46  & 0.46  \\
	\botrule
\end{tabular}}}
\end{tablehere}

\subsection{Experiments Result on Path Planning}
A simulation environment, as illustrated in Fig. \ref{path} and Fig. \ref{PP}, has been established to evaluate the efficacy of SEPSO in addressing the dynamic path planning problem. The size of the map is $366\ cm\times366\ cm$, which encompasses 6 dynamic obstacles (in black) and 2 static obstacles (in yellow). The velocity range for these dynamic obstacles is confined to the interval $(0, 5]\ cm/s$. 

Concerning the particles, the number of groups $G$ is set as 8, with each group comprising 170 individual particles. Each particle is represented by a 16-dimensional vector, corresponding to 8 waypoints. In addition, the parameters for the $Penalty$ function, PI, and AT are listed in Table \ref{params_path}. Note that these parameters and the particle configuration are handcrafted to strike a balance between the computational efficiency and quality of the generated path.

We first utilize the HSEF to determine the best hyper-parameters for the path planning problem, referred to as the evaluation phase of SEPSO. A total of 500 self-evolutions have been conducted, and the best hyper-parameters found are presented in Table \ref{hyper8path}. Following this, the application phase of SEPSO is performed, where the identified best hyper-parameters are integrated with the DTPSO to tackle the dynamic path planning problem, as the results illustrated in Fig. \ref{PP}.

To facilitate quantitative analysis and comparison, we maintain a fixed random seed (so that the behaviors of the obstacles and the start/target point are identical across the conducted experiments) and execute the simulation environment for 100 consecutive frames. Subsequently, we compare the results averaged over the 100 frames within the following algorithms: SEPSO, SEPSO(NoAT), SEPSO(NoPI), DTPSO, DPPSO, and PSO, as listed in Table \ref{ppresult}. Here, SEPSO refers to DTPSO with evolved hyper-parameters, while DTPSO designates DTPSO with unevolved hyper-parameters. Additionally, SEPSO(NoAT) represents SEPSO without auto truncation, and SEPSO(NoPI) denotes SEPSO without priori initialization. Note that for algorithms without auto truncation, the maximum iteration per frame is set as 30.

\begin{tablehere}
\tbl{Path Planning Performance Comparison\label{ppresult}}
{\begin{tabular}{llll}
    \toprule
    Algorithm    & \makecell[l]{Path\\Length}   & \makecell[l]{Time per\\Frame} & \makecell[l]{Iterations per\\Frame} \\
	\colrule
    SEPSO & 412.0   & 0.015 & 12.9 \\
    SEPSO(NoAT) & 408.3 & 0.036 & 30.0 \\
    SEPSO(NoPI) & 415.8 & 0.033 & 26.8 \\
    DTPSO & 413.9 & 0.036 & 30.0 \\
    DPPSO & 413.9 & 0.861 & 30.0 \\
    PSO   & 449.9 & 4.232 & 30.0 \\

	\botrule
\end{tabular}}
\end{tablehere}

Table \ref{ppresult} indicates that the SEPSO outperforms DTPSO, DPPSO, and PSO on all evaluation criteria, wherein the SEPSO achieves the shortest path length with the lowest planning time, providing valid corroboration for the efficacy of our approach. Upon comparing SEPSO with SEPSO(NoAT), it can be observed that the AT vastly improves the path planning speed (approximately 2 times more speed up) with a moderate level loss of optimality, which we contend is entirely acceptable for dynamic scenarios necessitating prompt planning and decision-making. Meanwhile, the PI effectively ameliorates the quality of the solution and concurrently reduces the number of iterations necessary for each planning phase, thus resulting in a shortened planning time.

\section{Conclusion and Future works}
This paper introduces a SEPSO algorithm to surmount the challenges of low computational efficiency and premature convergence inherent in the existing algorithms within the PSO family. These challenges are effectively resolved through the proposed TOF and HSEF. Comprehensive experimentation involving four benchmark optimization problems is conducted to substantiate the efficacy of the SEPSO. Harnessing the outstanding features of SEPSO, an efficacious approach for path planning in dynamic scenarios is formulated, wherein two succinct yet impactful mechanisms, namely the PI and AT, have been proposed and seamlessly integrated into the SEPSO to boost its real-time performance. It is noteworthy that the proposed approach successfully accomplishes the task of dynamic path planning in intricate environments containing both convex and non-convex obstacles with an impressive speed of 0.015 seconds per planning, amounting to roughly 67 plannings per second. We firmly contend that our proposed methodology could be deemed a potent solution for real-time path planning in dynamic scenarios. Future works could be directed towards: (i) the integration of environmental kinematics to further augment the rationality of the planned paths (ii) the application with unstructured environments (iii) the extension from 2D planning to 3D planning or from single agent system to multiple agents system.

\nonumsection{Acknowledgments} \noindent We acknowledge the support from the National Natural Science Foundation of China under Grant No. 62273230.

\appendix{Details of the Hardware Platform}

\begin{tablehere}
\tbl{Hardware Platform.\label{EP}}
{\resizebox{0.35\textwidth}{!}{
\begin{tabular}{@{}ll@{}}
\toprule
Component & Description \\
\colrule
CPU   & Intel Core i9-10850k  \\
GPU   & Nvidia RTX 2070 SUPER \\
RAM   & 32GB 3200MHz \\
System & Ubuntu 20.04.1 \\
\botrule
\end{tabular}}}
\end{tablehere}

\appendix{Hyper-parameters}

\begin{tablehere}
\tbl{Hyper-parameters for PSO. \label{hyper1}}
{
\resizebox{0.3\textwidth}{!}{
\begin{tabular}{lllll}
    \toprule
    $C1$    & $C2$   & $\omega_{init}$ & $\omega_{end}$ & $V_{limit}$ \\
	\colrule
    2     & 2     & 0.9     & 0.4   & 0.5  \\
	\botrule
\multicolumn{5}{l}{$V_{range}=V_{limit} \cdot X_{range}$}
\end{tabular}}}
\end{tablehere}

\begin{tablehere}
\tbl{Parameters of PI, AT, and $Penalty$ function. \ \ \ \ \ \ \ \ \  \label{params_path}}
{\begin{tabular}{lllll}
    \toprule
    $\alpha$  \ \ \ \ \ \   & $\beta$ \ \ \ \ \ \   & $\gamma$ \ \ \ \ \ \ & $\delta$ \ \ \ \ \ \ & $TW$ \\
	\colrule
    30     & 4     & 0.25    & 10  & 20  \\
	\botrule
\end{tabular}}
\end{tablehere}

\columnbreak

\begin{tablehere}
\tbl{Hyper-parameters for DTPSO, H-SEPSO, and DPPSO. \ \ \ \ \ \ \ \ \ \ \ \ \ \ \ \ \ \ \ \ \  \label{hyper8}}
{\begin{tabular}{@{}lllllll@{}}
    \toprule
    Group & $C1$    & $C2$    & $C3$    & $\omega_{init}$ & $\omega_{end}$ & $V_{limit}$ \\
	\colrule
    1     & 2     & 1     & 1     & 0.4   & 0.2   & 0.2 \\
    2     & 1     & 1     & 2     & 0.7   & 0.3  & 0.1 \\
    3     & 2     & 2     & 1     & 0.8   & 0.1  & 0.6 \\
    4     & 2     & 2     & 1     & 0.8   & 0.6  & 0.4 \\
    5     & 2     & 1     & 2     & 0.2   & 0.1   & 0.3 \\
    6     & 2     & 1     & 2     & 0.9   & 0.5  & 0.5 \\
    7     & 1     & 2     & 2     & 0.4   & 0.1  & 0.8 \\
    8     & 1     & 2     & 2     & 0.9   & 0.3  & 0.3 \\
	\botrule
\multicolumn{7}{l}{$V_{range}=V_{limit} \cdot X_{range}$}
\end{tabular}}
\end{tablehere}

\begin{tablehere}
\tbl{Hyper-parameters for L-SEPSO on Path Planning. \ \ \ \ \ \ \ \ \ \ \ \ \ \ \ \ \ \ \ \ \  \label{hyper8path}}
{\begin{tabular}{@{}lllllll@{}}
    \toprule
    Group & $C1$    & $C2$    & $C3$    & $\omega_{init}$ & $\omega_{end}$ & $V_{limit}$ \\
	\colrule
    1     & 1.53  & 1.29  & 1.34  & 0.48  & 0.19  & 0.35  \\
    2     & 1.72  & 1.53  & 1.34  & 0.73  & 0.28  & 0.32  \\
    3     & 1.34  & 1.42  & 1.33  & 0.48  & 0.21  & 0.62  \\
    4     & 1.76  & 1.60  & 1.21  & 0.47  & 0.30  & 0.63  \\
    5     & 1.68  & 1.27  & 1.25  & 0.73  & 0.36  & 0.41  \\
    6     & 1.66  & 1.54  & 1.54  & 0.39  & 0.16  & 0.45  \\
    7     & 1.57  & 1.48  & 1.75  & 0.56  & 0.34  & 0.38  \\
    8     & 1.31  & 1.71  & 1.23  & 0.36  & 0.25  & 0.50  \\
	\botrule
\multicolumn{7}{l}{$V_{range}=V_{limit} \cdot X_{range}$}
\end{tabular}}
\end{tablehere}

\bibliographystyle{ws-us}
\bibliography{SEPSO}

\noindent\includegraphics[width=1in]{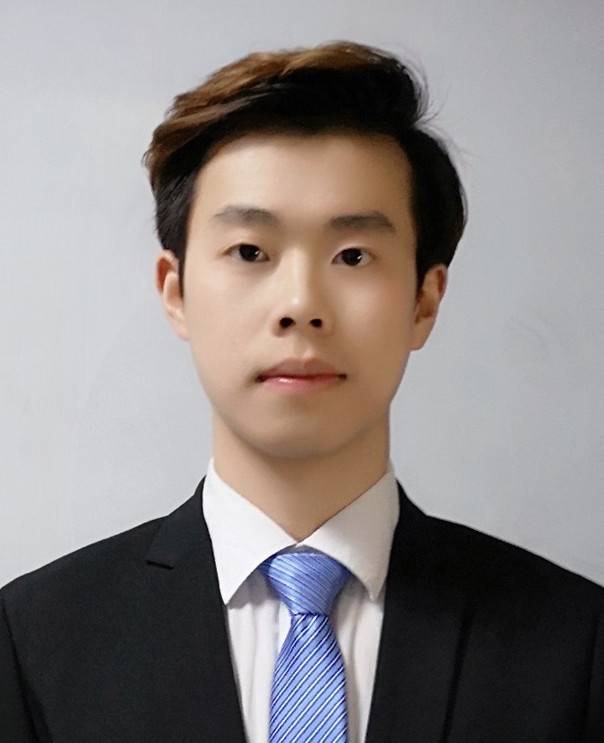}
{\bf Jinghao Xin} is a PhD candidate in the Department of Automation at Shanghai Jiao Tong University. He received his Bachelor degree in the Department of Automation at Ji Lin University. He is currently a visiting scholar in the School of Civil and Environmental Engineering at Nanyang Technological University. His research interests include Optimization, Deep Reinforcement Learning, and Path Planning. His most recent research can be found at https://github.com/XinJingHao.

\noindent\includegraphics[width=1in]{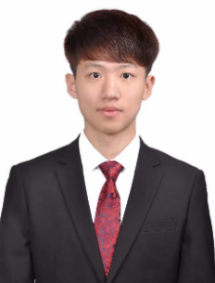}
{\bf Zhi Li} is a master candidate in the Department of Automation at Shanghai Jiao Tong University. His research interests include deep reinforcement learning, autonomous exploration and mapping for mobile robots, and SLAM.

\noindent\includegraphics[width=1in]{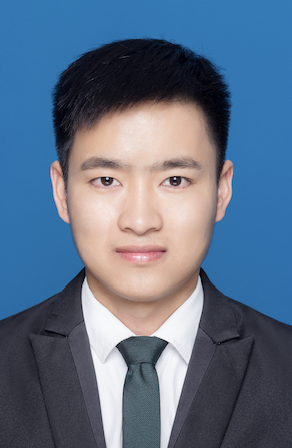}
{\bf Yang Zhang} is now a PhD candidate in the Department of Automation, Shanghai Jiao Tong University. He received a master's degree in Light Industry Technology and Engineering from South China University of Technology in 2021. His current research interests include power system planning, model predictive control and safe reinforcement learning.

\noindent\includegraphics[width=1in]{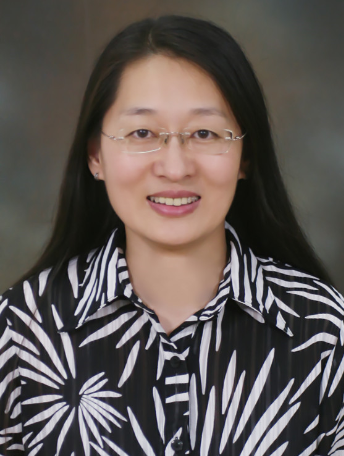}
{\bf Ning Li} (IEEE Member) attained her Bachelor's and Master's degrees from Qingdao University of Science and Technology, Qingdao, China, in 1996 and 1999 respectively. She further obtained her PhD degree from Shanghai Jiao Tong University, Shanghai, China, in 2002. Presently, she is a Professor within the Department of Automation at Shanghai Jiao Tong University, Shanghai, China. Her research interests encompass big data analysis, artificial intelligence, and modeling and control of complex systems.

\end{multicols}
\end{document}